\documentclass[10pt,twocolumn,letterpaper]{article}

\usepackage[pagenumbers]{cvpr} 
\usepackage{multirow}
\usepackage{booktabs}
\usepackage{adjustbox}

\usepackage{colortbl}

\definecolor{cvprblue}{rgb}{0.21,0.49,0.74}
\newcommand{\sigm}{\mathrm{sigm}}

\usepackage[pagebackref,breaklinks,colorlinks,allcolors=cvprblue]{hyperref}

\def\confName{CVPR}
\def\confYear{2026}

\newcommand{\algoname}{MambaFusion}

\title{\algoname: Adaptive State-Space Fusion for Multimodal \\ 3D Object Detection}

\author{
Venkatraman Narayanan$^{1}$  \quad
Bala Sai$^{1}$  \quad
Rahul Ahuja$^{1}$  \quad
Pratik Likhar$^{2}$  \\[2pt]
Varun Ravi Kumar$^{1}$  \quad
Senthil Yogamani$^{1}$  \\[6pt]
{\tt\small $^{1}$Automated Driving, Qualcomm Technologies, Inc} \\
{\tt\small $^{2}$Automated Driving, Qualcomm India Private Limited} 
}

\begin{document}
\maketitle
\begin{abstract}
Reliable 3D object detection is fundamental to autonomous driving, and multimodal fusion algorithms using cameras and LiDAR remain a persistent challenge. Cameras provide dense visual cues but ill posed depth; LiDAR provides a precise 3D structure but sparse coverage. Existing BEV-based fusion frameworks have made good progress, but they have difficulties including inefficient context modeling, spatially invariant fusion, and reasoning under uncertainty. We introduce \textbf{\algoname}, a unified multi-modal detection framework that achieves efficient, adaptive, and physically grounded 3D perception. \algoname~ interleaves selective state-space models (SSMs) with windowed transformers to propagate the global context in linear time while preserving local geometric fidelity. A multi-modal token alignment (MTA) module and reliability-aware fusion gates dynamically re-weight camera–LiDAR features based on spatial confidence and calibration consistency. Finally, a structure-conditioned diffusion head integrates graph-based reasoning with uncertainty-aware denoising, enforcing physical plausibility, and calibrated confidence. \algoname~ establishes new state-of-the-art performance on nuScenes benchmarks while operating with linear-time complexity. The framework demonstrates that coupling SSM-based efficiency with reliability-driven fusion yields robust, temporally stable, and interpretable 3D perception for real-world autonomous driving systems.
\end{abstract}    
\section{Introduction}
\label{sec:intro}

Robust 3D object detection is the cornerstone of autonomous perception. Cameras capture rich semantics but are unreliable for depth; LiDAR provides precise geometry yet remains sparse and costly. Achieving reliable perception, therefore, hinges on \emph{how effectively these modalities are fused} into a unified spatial understanding of the scene.

Recent bird’s-eye-view (BEV) frameworks~\cite{bevfusion, transfusion,cmt,bevformer, borse2023x, schramm2024bevcar} have transformed multi-sensor fusion by projecting heterogeneous features into a common geometric space.  
However, despite steady progress, four long-standing challenges continue to limit performance.  
\textbf{(1) Inefficient context modeling:} transformer-based 3D encoders offer global awareness but scale quadratically, making long-range and temporal reasoning computationally expensive.  
\textbf{(2) Spatially invariant fusion:} most systems fuse modalities with fixed or global weights, ignoring that sensor reliability varies with distance, occlusion, and calibration drift.  
\textbf{(3) Lack of physical reasoning:} current detectors predict object confidence purely from feature activations, without enforcing geometric plausibility or structural consistency.  
\textbf{(4) Temporal instability:} detections often fluctuate across frames due to the absence of cross-time feature constraints.

We introduce \textbf{\algoname}, a unified, end-to-end architecture that bridges these gaps through efficient state-space modeling, adaptive reliability-aware fusion, and structure-conditioned reasoning.  
First, a \emph{hybrid LiDAR encoder} interleaves Mamba state-space blocks with windowed attention, enabling linear-time propagation of global context while preserving fine local geometry.  
Second, an \emph{adaptive multi-modal fusion decoder} aligns camera and LiDAR features via a learnable token alignment (MTA) module, followed by spatial reliability gates that dynamically adjust sensor contributions based on point density, multi-view consistency, and projection confidence.  
Third, a \emph{structure-conditioned diffusion head} refines detections through graph-based reasoning and uncertainty-guided denoising, enforcing physical plausibility and calibrated confidence.  
Finally, a temporal self-distillation objective stabilizes features across frames, yielding temporally consistent and trustworthy perception.

Our contributions are summarized as follows:
\begin{itemize}[leftmargin=*,noitemsep,topsep=3pt]
    \item \textbf{Hybrid linear-time LiDAR encoder:} interleaving Mamba state-space blocks and windowed transformers achieves efficient global–local modeling without quadratic attention overhead.
    \item \textbf{Adaptive reliability-aware fusion:} a learnable alignment and gating mechanism dynamically reweights camera–LiDAR features based on spatial confidence and sensor reliability.
    \item \textbf{Structure-conditioned diffusion refinement:} integrates graph reasoning and diffusion-based denoising to enforce geometric plausibility and calibrated uncertainty.
    \item \textbf{Unified, temporally stable perception:} \algoname~ delivers state-of-the-art performance on nuScenes dataset with superior robustness to calibration noise, sparsity, and dynamic scenes.
\end{itemize}

\algoname~ demonstrates that coupling selective state-space models with reliability-driven fusion yields an efficient, interpretable, and physically grounded approach to multi-sensor 3D perception.
\section{Related Work}

\subsection{Camera-based 3D Object Detection}
Camera-only 3D detection methods typically transform multi-view features into bird's-eye-view (BEV) representations, enabling spatial reasoning. 
Lift–Splat–Shoot (LSS)~\cite{philion2020lift} first predicted per-pixel depth distributions to lift image features into BEV space. 
BEVDet~\cite{huang2021bevdet} and BEVDepth~\cite{li2023bevdepth} improved geometric accuracy through explicit depth supervision, while temporal extensions such as BEVDet4D~\cite{huang2022bevdet4d} and SOLOFusion~\cite{park2023solofusion} exploited temporal aggregation for motion understanding. 
Query-based approaches (e.g., DETR3D~\cite{wang2021detr3d}, PETR~\cite{liu2022petr}) introduced learnable 3D queries projected into image space for end-to-end detection. 
BEVFormer~\cite{li2022bevformer} unified these ideas with deformable spatial cross-attention and temporal transformers, setting the foundation for camera-based BEV feature extraction. 
Our work adopts BEVFormer's multi-view encoder but replaces quadratic temporal attention with a linear-time state-space model to improve scalability.

\subsection{LiDAR-based 3D Object Detection}
LiDAR sensors provide explicit geometry but remain computationally expensive. 
VoxelNet~\cite{zhou2018voxelnet} pioneered voxel-based encoding; SECOND~\cite{yan2018second} introduced sparse 3D convolutions for efficiency; and CenterPoint~\cite{yin2021centerpoint} popularized center-based detection. 
Recent transformer-based backbones, such as VoTr~\cite{mao2021votr} and DSVT~\cite{wang2023dsvt}, have improved global context modeling but suffer from quadratic complexity with respect to voxel count. 
This motivates the need for alternative sequence models that can retain global awareness while scaling linearly with scene size—a capability our Mamba-based LiDAR encoder provides.

\subsection{Efficient Sequence Models for Vision}
State-space models (SSMs) have recently emerged as efficient alternatives to traditional models, such as transformers. 
Mamba~\cite{gu2023mamba} introduced selective SSMs that achieve input-dependent long-range modeling with linear complexity, inspiring variants in vision tasks~\cite{zhu2024vim,liu2024vmamba}. 
PointMamba~\cite{liang2024pointmamba} and PointSSM~\cite{zhang2024pointssm} extended this idea to point clouds, demonstrating that SSMs can capture geometric continuity effectively. 
Hybrid designs~\cite{hatamizadeh2025mambavision} that interleave SSMs for global propagation with local attention for detailed reasoning show strong potential for structured data. 
{\algoname~} builds directly on this paradigm, alternating SSM and windowed attention blocks to jointly model long-range temporal dependencies and local 3D structure.

\subsection{Windowed Attention Mechanisms}
Windowed attention reduces the quadratic cost of global transformers while preserving spatial locality. 
Swin Transformer~\cite{liu2021swin} introduced shifted windows for efficient image modeling, later extended to 3D through stratified or sliding-window designs~\cite{sun2022swformer,lai2022stratified}. 
In {\algoname~}, windowed attention complements SSM propagation by refining local geometry and ensuring accurate spatial alignment in BEV space.

\subsection{Multi-Modal Fusion for 3D Detection}
Integrating complementary sensors remains a significant challenge in autonomous perception. 
Early fusion methods (PointPainting~\cite{vora2020pointpainting}, MVX-Net~\cite{sindagi2019mvx}) combined image semantics with LiDAR points directly, while BEV-space fusion later emerged as the dominant approach. 
BEVFusion~\cite{liu2023bevfusion} unified camera and LiDAR features in a shared BEV plane, TransFusion~\cite{bai2022transfusion} used transformer decoders for soft association, and CMT~\cite{yan2023cmt} simplified fusion through token interaction. 
However, these methods typically apply fixed or globally shared fusion weights, thereby ignoring spatially varying reliability and calibration drift. 
Our \textbf{Multi-Modal Token Alignment (MTA)} explicitly corrects calibration offsets. At the same time, the adaptive fusion gate dynamically modulates sensor contribution based on reliability signals such as point density and cross-view consistency.

\subsection{Uncertainty and Calibration in Perception}
Accurate confidence estimation is essential for robust fusion. 
Aleatoric and epistemic uncertainty modeling~\cite{kendall2017uncertainties} improves reliability, and evidential regularization~\cite{sensoy2018evidential} encourages calibrated predictions. 
In multi-sensor systems, spatial misalignment and synchronization errors degrade fusion performance. Recent works address this issue through pose correction or probabilistic alignment. 
{\algoname~} integrates uncertainty-aware weighting directly into the fusion process, using inverse-variance fusion to downweight unreliable regions and ensure stable multi-modal aggregation.

\subsection{Spatial Reasoning and Diffusion-based Refinement}
Spatial context helps resolve ambiguous detections. 
Graph-based reasoning, as in PointGNN~\cite{shi2020pointgnn}, models object relationships for consistency, while diffusion-based detectors~\cite{chen2023diffusiondet,chen2024diffubox} formulate detection as iterative denoising for confidence refinement. 
We unify these ideas by coupling graph-based spatial reasoning with a \textbf{structure-conditioned diffusion module} that refines proposal confidences under geometric and contextual guidance. 
This combination enables both physically plausible detections and calibrated confidence estimation.

\section{Methodology}
\label{sec:method}

\begin{figure*}[t]
    \centering
    \includegraphics[width=0.89\linewidth]{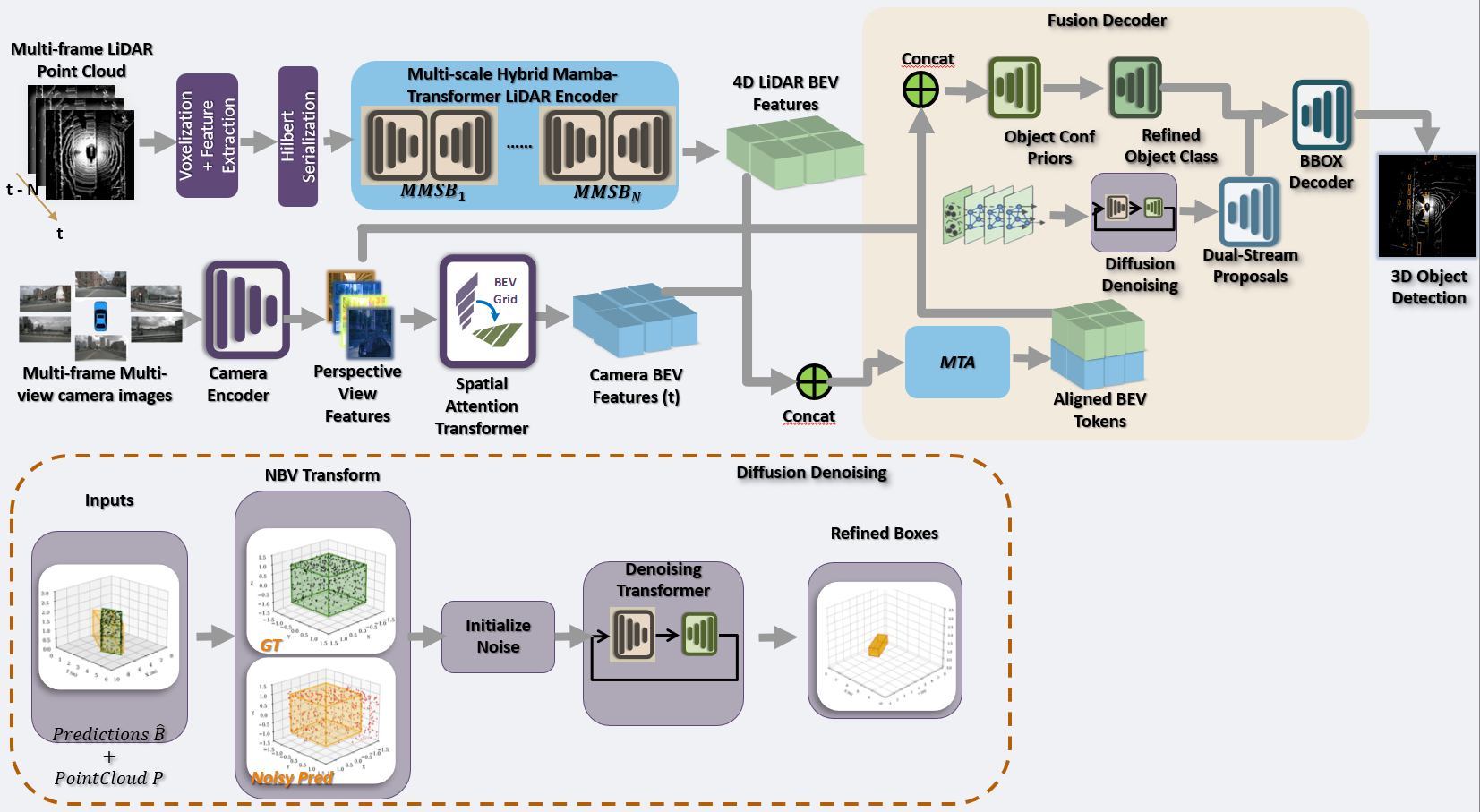}
    \caption{\textbf{Architecture Overview.} {\algoname~} features: (1) multi-frame camera and LiDAR encoding with spatiotemporal Transformers and hybrid Mamba SSM/Transformer blocks, (2) Multi-Modal Token Alignment (MTA) for robust cross-modal spatial calibration, (3) uncertainty-aware adaptive fusion via bidirectional attention and spatial reliability gating, (4) dual-stream proposal generation and geometry-aware graph reasoning, (5) structure-conditioned diffusion for confidence refinement, and (6) temporal self-distillation for prediction stability. All components are jointly optimized for robust multi-modal 3D object detection.}
    \label{fig:architecture}
\end{figure*}

We propose \textbf{\algoname~}, a unified, end-to-end framework for multi-modal 3D object detection that addresses three core challenges in sensor fusion: (i) efficiently encoding long-range and fine-grained structure in sparse 3D data, (ii) dynamically weighting sensor contributions according to spatial reliability and contextual cues, and (iii) explicitly modeling physical plausibility and uncertainty in detection outcomes. 
Existing fusion methods~\cite{liu2023bevfusion,yan2023cmt,bai2022transfusion,yang2023deepinteraction} typically employ spatially uniform fusion or deterministic weighting, limiting robustness to miscalibration, sparsity, and ambiguous object geometry. 
An overview of the whole pipeline is illustrated in Fig.~\ref{fig:architecture}.

\subsection{Multi-Scale Hybrid LiDAR Encoding}

Sparse convolutions offer efficiency but limited receptive fields, while pure attention achieves global context at quadratic cost. 
We resolve this trade-off via a hybrid LiDAR encoder that alternates \textbf{Mamba State-Space Model (SSM)} blocks~\cite{mamba,liang2024pointmamba} for global context propagation and \textbf{windowed Transformers}~\cite{liu2021swin} for local spatial reasoning. 
Raw LiDAR points are voxelized, serialized along a Hilbert space-filling curve to preserve spatial adjacency, and processed through multiple spatial scales with alternating SSM and attention blocks. 
SSM layers model long-range dependencies in linear time, while windowed attention captures fine structural detail. 
Cross-scale aggregation and residual links generate multi-scale BEV features that balance global awareness and local precision. 
We denote the resulting LiDAR BEV representation as $Q_{\mathrm{L}}^{\mathrm{BEV}}$.

\subsection{Spatiotemporal Camera BEV Aggregation}

Multi-view images are encoded using a shared backbone with feature pyramid network (FPN) features. 
Following~\cite{li2022bevformer}, learnable BEV queries attend to multi-scale image features via deformable cross-attention, producing aligned BEV tokens $Q_{\mathrm{C}}^{\mathrm{BEV}}$. 
We denote by $B_t$ the BEV token grid at time $t$ (concatenated camera BEV tokens after projection). 
Temporal \textbf{Mamba blocks} operate on $\{B_t\}$ to capture long-term dependencies across frames with linear complexity in time and token count. 
In addition to the LiDAR-stream SSM, we apply Mamba over $\{B_t\}$ for linear-time temporal aggregation, replacing quadratic temporal attention and enabling efficient modeling of motion and scene continuity.

\subsection{Multi-Modal Token Alignment (MTA)}

Extrinsic calibration between cameras and LiDAR is susceptible to degradation during vehicle operation due to thermal expansion, vibration, and mounting stress~\cite{lccnet2021,calibnet2018}. 
Studies have shown that calibration can drift by 0.5-2° over time in deployed systems~\cite{automatic_calibration_rss2013,thermal_calibration2019}, necessitating regular recalibration.
To maintain robust fusion under such real-world conditions, we introduce a lightweight \textbf{Multi-Modal Token Alignment (MTA)} module that learns to correct residual calibration and synchronization errors between modalities.

MTA learns positional offsets between camera and LiDAR BEV tokens to align them in a shared coordinate frame:
\vspace{-2pt}
\begin{equation}
\begin{aligned}
\Delta P &= f_\phi(Q_{\mathrm{C}}^{\mathrm{BEV}}, Q_{\mathrm{L}}^{\mathrm{BEV}}), \\
Q_{\mathrm{C}}' &= Q_{\mathrm{C}}^{\mathrm{BEV}} + \Delta P, \quad
Q_{\mathrm{L}}' = Q_{\mathrm{L}}^{\mathrm{BEV}} - \Delta P,
\end{aligned}
\end{equation}
\vspace{-2pt}
where $f_\phi$ is a lightweight cross-attention network that estimates spatial offsets from the input BEV features.
This learned alignment mitigates minor extrinsic drift and improves cross-modal correspondence, enhancing fusion robustness to calibration variations encountered in practice.

\subsection{Adaptive Fusion with Uncertainty Modeling}

We fuse $Q_{\mathrm{C}}'$ and $Q_{\mathrm{L}}'$ through bidirectional multi-head attention (MHA), allowing each modality to incorporate complementary cues:
\begin{align}
Q_{\mathrm{C}}^{\mathrm{att}} &= \text{MHA}(Q_{\mathrm{C}}', K=Q_{\mathrm{L}}', V=Q_{\mathrm{L}}'), \\
Q_{\mathrm{L}}^{\mathrm{att}} &= \text{MHA}(Q_{\mathrm{L}}', K=Q_{\mathrm{C}}', V=Q_{\mathrm{C}}').
\end{align}

\paragraph{Spatial reliability gating.}
For each BEV cell $(x,y)$, we derive a descriptor $\mathbf{g}(x,y)$ combining point density $\rho_L$, depth variance $\tau_C$, occlusion score $o$, multi-view consistency $\phi$, and ego distance $d$. 
A small MLP predicts a fusion gate:
\vspace{-2pt}
\begin{align}
g(x,y) = \sigm\!\big(\Phi_{\text{gate}}([\mathbf{g}(x,y), Q_{\mathrm{C}}^{\mathrm{att}}(x,y), Q_{\mathrm{L}}^{\mathrm{att}}(x,y)])\big),
\end{align}
\vspace{-2pt}
where $\sigm(\cdot)$ denotes the sigmoid. 
This gate adaptively weights sensors across the BEV space based on reliability indicators.

\paragraph{Uncertainty-aware weighting.}
Each modality predicts a per-cell log-variance map $\log\sigma_m^2(x,y)$ via lightweight heads $\Psi_m$. 
We perform inverse-variance fusion:
\vspace{-2pt}
\begin{align}
Q_{\text{fused}}(x,y) =
\frac{g\,Q_{\mathrm{C}}^{\mathrm{att}}/\sigma_C^2 + (1-g)\,Q_{\mathrm{L}}^{\mathrm{att}}/\sigma_L^2}
{g/\sigma_C^2 + (1-g)/\sigma_L^2 + \epsilon},
\end{align}
\vspace{-2pt}
with $\epsilon=10^{-6}$ for stability and $\sigma_m^2\!\in\![\epsilon,\sigma_{\max}^2]$. 
We stop the gradient on $\sigma_m^2$ for the first two epochs to stabilize training. 
High-uncertainty regions automatically contribute less, and random modality dropout encourages robustness under partial sensor loss.

\subsection{Dual-Stream Proposal Generation and Spatial Reasoning}

From fused BEV features, we form proposals by combining modality-specific heatmaps:
\vspace{-2pt}
\begin{align}
H_{\text{fused}} = \beta_C H_{\text{cam}} + \beta_L H_{\text{lidar}},
\end{align}
\vspace{-2pt}
selecting global top-$K$ peaks (we use $K{=}900$) as initial boxes $\mathbf{b}_i$ with features $\mathbf{h}_i$ and confidences $c_i^{\text{init}}$.

To enforce physical plausibility, we construct a spatial graph $\mathcal{G}=(\mathcal{V},\mathcal{E})$ over proposals, connecting each node to its $k$ nearest neighbors within $r{=}4$ m ($k{=}16$ unless noted). 
Edges are recomputed each iteration to reflect updated embeddings. 
Each node augments its features with structural descriptors (class–size consistency, ground-plane offset, LiDAR support, and confidence). 
Message passing refines proposal embeddings:
\vspace{-2pt}
\begin{align}
\mathbf{m}_{i\to j}^{(t)} &= \text{MLP}_{\text{msg}}([\mathbf{h}_i^{(t)},\mathbf{h}_j^{(t)},\mathbf{g}_i,\mathbf{g}_j,\mathbf{e}_{ij}]), \\
\mathbf{h}_i^{(t+1)} &= \text{MLP}_{\text{upd}}\!\Big(\mathbf{h}_i^{(t)} + \sum_{j\in\mathcal{N}(i)} \mathbf{m}_{j\to i}^{(t)}\Big).
\end{align}
\vspace{-2pt}
This reasoning suppresses implausible configurations such as overlapping or floating objects.

\subsection{Structure-Conditioned Diffusion Refinement (GCD)}

We refine proposal confidences through a diffusion-based refinement process conditioned on spatial reliability. 
Each proposal feature–confidence pair $\mathbf{z}_i=[\mathbf{h}_i,c_i]$ undergoes a conditional denoising process:
\vspace{-2pt}
\begin{align}
q(\mathbf{z}_t|\mathbf{z}_0)=\mathcal{N}(\mathbf{z}_t;\sqrt{\bar\alpha_t}\mathbf{z}_0,(1-\bar\alpha_t)\mathbf{I}),
\end{align}
\vspace{-2pt}
where the diffusion noise level depends on confidence $u_i\!\in\![0,1]$ estimated from LiDAR support, ground proximity, and local camera cues:
\vspace{-2pt}
\begin{equation}
\begin{aligned}
u_i &= \sigm\!\big(w_1\,\hat N_{\text{pts}}(i)
      - w_2\,|z_i - z_{\text{ground}}|
      - w_3\,\tau_C^{\text{local}}(i)\big), \\
\sigma_i &= \sigma_0 + \alpha(1 - u_i).
\end{aligned}
\end{equation}
\vspace{-2pt}
The denoiser $\epsilon_\theta(\mathbf{z}_t,t,\mathbf{g}_i,\mathbf{F}_{\text{local}})$ is trained with
\vspace{-2pt}
\begin{align}
\mathcal{L}_{\text{diff}} = 
\mathbb{E}_{t,\mathbf{z}_0,\epsilon}\!
\big[\|\epsilon - \epsilon_\theta(\mathbf{z}_t,t,\mathbf{g}_i,\mathbf{F}_{\text{local}})\|^2\big].
\end{align}
\vspace{-2pt}
During inference, we use $S{=}3$ reverse steps, which empirically balance speed and accuracy, to obtain refined confidences consistent with multi-sensor context.

\paragraph{Spatial Priors Discussion.}
Although \algoname~ integrates multiple modules that utilize structural information, each operates at a distinct stage of the perception pipeline and contributes a unique supervisory signal. 
The adaptive fusion gate modulates sensor reliability within BEV space, the proposal-level reasoning network enforces physical plausibility among nearby objects, the diffusion module refines confidence according to spatial uncertainty, and the consistency loss provides a weak global prior during training. 
To prevent over-constraining the optimization, $\mathcal{L}_{\text{geo}}$ is assigned a small weight (0.1–0.3), ensuring that structural priors act as stabilizing cues rather than dominant objectives. 
This multi-level treatment of spatial context improves robustness in sparse or ambiguous scenes while maintaining flexibility to learn complex semantics.

\subsection{Temporal Self-Distillation (TSD)}

We introduce \textbf{Temporal Self-Distillation} to stabilize predictions across frames. 
Given BEV embeddings $B_t$ and Mamba state $h_t$, the network predicts the next embedding $\hat{B}_{t+1}=f_{\text{pred}}(B_t,h_t)$. 
A self-distillation loss enforces temporal coherence:
\vspace{-2pt}
\begin{align}
\mathcal{L}_{\text{temp}} = \| \hat{B}_{t+1} - \text{stopgrad}(B_{t+1}) \|_1.
\end{align}
\vspace{-2pt}
This enables smooth temporal evolution without explicit motion labels and enhances detection stability across varying clip lengths and strides.

\subsection{End-to-End Optimization}

A transformer decoder~\cite{carion2020detr,zhu2020deformable} processes refined proposals to output final boxes $\hat{\mathbf{b}}_i$, confidences $\hat{c}_i$, and class logits $\hat{y}_i$. 
All components—from LiDAR and camera encoders to gating, uncertainty heads, GNN reasoning, diffusion refinement, and detection head—are trained jointly. 
The total loss is
\vspace{-2pt}
\begin{align}
\mathcal{L} =
\lambda_{\text{cls}}\mathcal{L}_{\text{cls}}
+ \lambda_{\text{reg}}\mathcal{L}_{\text{reg}}
+ \lambda_{\text{iou}}\mathcal{L}_{\text{iou}}
+ \lambda_{\text{unc}}\mathcal{L}_{\text{unc}} \nonumber \\
+ \lambda_{\text{geo}}\mathcal{L}_{\text{geo}} 
+ \lambda_{\text{diff}}\mathcal{L}_{\text{diff}}
+ \lambda_{\text{temp}}\mathcal{L}_{\text{temp}}.
\end{align}
\vspace{-2pt}
Here $\mathcal{L}_{\text{unc}} = \|\log\sigma_C^2\|_1 + \|\log\sigma_L^2\|_1$ regularizes uncertainty maps, $\mathcal{L}_{\text{geo}}$ penalizes structural constraint violations, and $\mathcal{L}_{\text{temp}}$ enforces temporal consistency.

\paragraph{Design Rationale.}
\algoname~ introduces seven supervised components, each addressing a distinct limitation observed in prior BEV fusion systems and collectively forming a minimal, non-redundant set. 
The classification, regression, and IoU losses handle the core detection objectives by supervising category, localization, and overlap quality. 
The uncertainty and diffusion losses regulate confidence estimation and epistemic uncertainty, enabling reliable sensor weighting and consistent fusion across modalities. 
The structural and temporal losses promote physical plausibility and temporal coherence, preventing implausible object configurations and reducing prediction flicker across frames. 
These objectives operate at complementary levels—object, fusion, and scene—and provide non-overlapping gradients that improve optimization stability. 
Ablation studies on nuScenes (Table~\ref{tab:nus_core_ablation}) show that removing any term results in a measurable drop in NDS or mAP, confirming that the seven-term formulation is compact and necessary for robust end-to-end performance.

\subsection{Complexity and Discussion}

\algoname~ achieves \(\mathcal{O}(TN)\) temporal complexity for $T$ frames and $N=H_{\text{bev}}\!\times W_{\text{bev}}$ BEV tokens (e.g., $200{\times}200$ in nuScenes), compared to \(\mathcal{O}(TN^2)\) for attention-based temporal models. 
All runtime metrics are measured on an NVIDIA A100 (80\, GB) with $900{\times}1600$ input resolution, a $200{\times}200$ BEV grid, and clip length 8. 
This linear-time formulation significantly improves scalability while maintaining accuracy. 
The uncertainty-aware fusion and diffusion refinement enhance robustness in sparse or ambiguous regions, whereas MTA and TSD ensure calibration and temporal stability. 
Together, these components yield a \textit{spatiotemporally adaptive, uncertainty-aware, and fully differentiable} BEV fusion framework.

\section{Experiments}

We conduct comprehensive experiments on two large-scale autonomous driving benchmarks to validate our framework, followed by thorough ablation studies and qualitative analysis.

\subsection{Experimental Setup}

\textbf{Datasets and Metrics.} We evaluate on nuScenes~\cite{caesar2020nuscenes} and Argoverse 2~\cite{wilson2023argoverse2}. nuScenes contains 1,000 driving scenes collected in Boston and Singapore, split into 700 training, 150 validation, and 150 test scenes, with 6 cameras and a 32-beam LiDAR covering 10 object categories. Argoverse 2 comprises 700 training sequences and 150 validation sequences, collected across multiple U.S. cities, utilizing 7 ring cameras and 2 stereo cameras, alongside a 32-beam LiDAR, for 3 categories. For nuScenes, we report NDS (a weighted combination of mAP and attribute errors, including translation, scale, orientation, velocity, and heading) and mAP at an IoU threshold of 0.5. For Argoverse 2, we report AP at IoU thresholds of 0.5 for 3D bounding boxes.

\textbf{Implementation.} Our camera branch uses Swin-T~\cite{liu2021swin} as the backbone with 6 encoder layers and 8 attention heads. The BEV space is discretized into a $200 \times 200$ grid with a resolution of 0.5m, covering a $[-50m, 50m]^2$ region around the ego-vehicle. For LiDAR processing, we use voxel sizes of $(0.3\text {m})^3$ for nuScenes and $(0.32\text {m})^3$ for Argoverse 2. The hybrid LiDAR encoder employs $L=4$ scales with downsampling factors $\{1, 2, 4, 8\}$ and $\{2, 2, 4, 2\}$ blocks per scale, using windowed attention with window size $w=7$. The fusion decoder has 6 layers and applies modality dropout with $p_{drop}=0.15$ during training. Our detection head uses $K=300$ object queries refined through $L=6$ decoder layers, while the geometry-aware GNN performs $M=3$ message-passing iterations with neighborhood radius $r_{neighbor}=5m$ and IoU threshold $\tau_{iou}=0.1$. All BEV features use $C=256$ dimensions. For confidence refinement, we use three geometry-conditioned diffusion steps with $\sigma_{min}=0.01$ and $\sigma_{max}=1.0$, extracting local context within a $2\times$ box dimension.

\textbf{Training and Inference.} We follow standard training protocols from BEVFormer~\cite{li2022bevformer} and BEVFusion~\cite{liu2023bevfusion}, training end-to-end with AdamW optimizer ($lr=2 \times 10^{-4}$, cosine annealing) on 8 A100 GPUs for 20 epochs (nuScenes) / 24 epochs (Argoverse 2). Loss weights: $\lambda_{cls}=2.0$, $\lambda_{reg}=0.25$, $\lambda_{iou}=2.0$, $\lambda_{unc}=0.1$, $\lambda_{geo}=0.2$, $\lambda_{diff}=0.15$, $\lambda_{temp}=0.1$ (auxiliary losses applied after 5-epoch warmup). For inference, we apply test-time augmentation with horizontal flipping and use Weighted Boxes Fusion~\cite{solovyev2021wbf} to merge predictions from different augmentations.

\subsection{Main Results}

\paragraph{nuScenes Validation.}
Table~\ref{tab:nuscenes_val} compares our method with recent multi-modal detectors on the nuScenes validation set. 
\textbf{MambaFusion} achieves \textbf{77.9 NDS} and \textbf{74.9 mAP}, establishing new state-of-the-art performance among methods using the Swin-T backbone. 
Compared to the strongest Swin-T baseline (BEVFusion4D: 73.5 NDS, 72.0 mAP), our method improves by \textbf{+4.4 NDS} and \textbf{+2.9 mAP}, demonstrating the effectiveness of our hybrid temporal encoding, adaptive fusion, and geometry-conditioned refinement (Section~\ref{sec:ablation}).

\paragraph{nuScenes Test Set.}
Table~\ref{tab:nuscenes_test} presents results on the official nuScenes test set. 
\textbf{MambaFusion} achieves \textbf{77.2 NDS} and \textbf{74.7 mAP} without ensemble or test-time augmentation, surpassing MV2DFusion (76.7 NDS, 74.5 mAP) which uses a significantly larger ConvNeXt-L backbone.
Our method maintains consistently strong performance across all error metrics: mATE (0.248m), mASE (0.232), mAOE (0.291 rad), mAVE (0.221 m/s), and mAAE (0.118), demonstrating balanced improvements in localization, scale, orientation, velocity, and attribute estimation.
This validates the generalization capability of our spatiotemporally adaptive fusion mechanism and confirms that the proposed components transfer effectively from validation to test scenarios.

\begin{table*}[t]
\centering
\caption{\textbf{Comparison on nuScenes validation set.} 
Methods are grouped by encoder architecture and sorted by increasing mAP. 
$\dagger$: uses CBGS training strategy.}
\label{tab:nuscenes_val}
\begin{small}
\begin{tabular}{l|c|c|c|cc}
\hline
Method & Modality & Image Backbone & LiDAR Backbone & NDS$\uparrow$ & mAP$\uparrow$ \\
\hline
\multicolumn{6}{c}{\cellcolor{blue!10}\textit{ResNet + VoxelNet}} \\
FUTR3D & C+L & ResNet-101 & VoxelNet & 68.0 & 64.2 \\
TransFusion & C+L & ResNet-50 & VoxelNet & 71.3 & 67.5 \\
DeepInteraction & C+L & ResNet-50 & VoxelNet & 72.6 & 69.9 \\
\hline
\multicolumn{6}{c}{\cellcolor{green!10}\textit{Swin-T + VoxelNet}} \\
BEVFusion & C+L & Swin-T & VoxelNet & 71.4 & 68.5 \\
BEVFusion & C+L & Swin-T & VoxelNet & 72.1 & 69.6 \\
BEVFusion4D-S & C+L & Swin-T & VoxelNet & 72.9 & 70.9 \\
SparseFusion & C+L & Swin-T & VoxelNet & 73.1 & 71.0 \\
EA-LSS & C+L & Swin-T & VoxelNet & 73.1 & 71.2 \\
BEVFusion4D & C+L & Swin-T & VoxelNet & 73.5 & 72.0 \\
\hline
\multicolumn{6}{c}{\cellcolor{purple!10}\textit{V2-99 + VoxelNet}} \\
FusionFormer-S$\dagger$ & C+L & V2-99 & VoxelNet & 73.2 & 70.0 \\
CMT & C+L & V2-99 & VoxelNet & 72.9 & 70.3 \\
SparseLIF-S & C+L & V2-99 & VoxelNet & 74.6 & 71.2 \\
FusionFormer$\dagger$ & C+L & V2-99 & VoxelNet & 74.1 & 71.4 \\
SparseLIF & C+L & V2-99 & VoxelNet & 77.5 & 74.7 \\
\hline
\multicolumn{6}{c}{\cellcolor{orange!10}\textit{Other Architectures}} \\
AutoAlignV2 & C+L & CSPNet & VoxelNet & 71.2 & 67.1 \\
MV2DFusion & C+L & ConvNeXt-L & VoxelNet & 75.4 & 73.9 \\
\hline
\rowcolor{yellow!20}
\textbf{MambaFusion (Ours)} & C+L & Swin-T & Hybrid SSM-Attn & \textbf{77.9} & \textbf{74.9} \\
\hline
\end{tabular}
\end{small}
\end{table*}

\begin{table*}[t]
\centering
\footnotesize
\setlength{\tabcolsep}{2.5pt}
\renewcommand{\arraystretch}{0.88}
\caption{\textbf{Comparison on nuScenes test set.} 
Methods sorted by increasing mAP within each group. 
-E indicates methods with model ensemble and TTA.}
\label{tab:nuscenes_test}
\begin{tabular}{l|c|cc|ccccc}
\toprule
Method & Mod. & NDS$\uparrow$ & mAP$\uparrow$ & mATE$\downarrow$ & mASE$\downarrow$ & mAOE$\downarrow$ & mAVE$\downarrow$ & mAAE$\downarrow$ \\
\midrule
\multicolumn{9}{c}{\cellcolor{blue!10}\textit{Single Model}} \\
\midrule
PointPainting & L & 61.0 & 54.1 & 0.380 & 0.260 & 0.541 & 0.293 & 0.131 \\
MVP & C+L & 70.5 & 66.4 & 0.263 & 0.238 & 0.321 & 0.313 & 0.134 \\
UVTR & C+L & 71.1 & 67.1 & 0.306 & 0.245 & 0.351 & 0.225 & 0.124 \\
TransFusion & C+L & 71.7 & 68.9 & 0.259 & 0.243 & 0.359 & 0.288 & 0.127 \\
BEVFusion & C+L & 72.9 & 70.2 & 0.261 & 0.239 & 0.329 & 0.260 & 0.134 \\
DeepInteraction & C+L & 73.4 & 70.8 & 0.257 & 0.240 & 0.325 & 0.245 & 0.128 \\
BEVFusion & C+L & 73.3 & 71.3 & 0.250 & 0.240 & 0.359 & 0.254 & 0.132 \\
CMT & C+L & 74.1 & 72.0 & 0.279 & 0.235 & 0.308 & 0.259 & 0.112 \\
EA-LSS & C+L & 74.4 & 72.2 & 0.247 & 0.237 & 0.304 & 0.250 & 0.133 \\
FusionFormer & C+L & 75.1 & 72.6 & 0.267 & 0.236 & 0.286 & 0.225 & 0.105 \\
BEVFusion4D & C+L & 74.7 & 73.3 & 0.252 & 0.237 & 0.315 & 0.248 & 0.127 \\
MV2DFusion & C+L & 76.7 & 74.5 & 0.245 & 0.229 & 0.269 & 0.199 & 0.115 \\
\midrule
\rowcolor{yellow!20}
\textbf{MambaFusion} & C+L & \textbf{77.2} & \textbf{74.7} & \textbf{0.248} & \textbf{0.232} & \textbf{0.291} & \textbf{0.221} & \textbf{0.118} \\
\midrule
\multicolumn{9}{c}{\cellcolor{green!10}\textit{With Ensemble \& TTA}} \\
\midrule
BEVFusion-E & C+L & 76.1 & 75.0 & 0.242 & 0.227 & 0.320 & 0.222 & 0.130 \\
CMT-E & C+L & 77.0 & 75.3 & 0.233 & 0.220 & 0.271 & 0.212 & 0.127 \\
EA-LSS-E & C+L & 77.6 & 76.6 & 0.234 & 0.228 & 0.278 & 0.204 & 0.124 \\
BEVFusion4D-E & C+L & 77.2 & 76.8 & 0.229 & 0.229 & 0.302 & 0.225 & 0.135 \\
MV2DFusion-E & C+L & 78.8 & 77.9 & 0.237 & 0.226 & 0.247 & 0.192 & 0.119 \\
\bottomrule
\end{tabular}
\end{table*}

\subsection{Ablation Studies}
\label{sec:ablation}

We conduct extensive ablations on the \textbf{nuScenes val} split to evaluate the contribution of each proposed component. 
All experiments use the same backbone, BEV grid size, and training schedule unless stated otherwise. 
We report official nuScenes metrics: NDS, mAP, mATE (m), mASE, mAOE (rad), and mAVE (m/s), along with efficiency (FPS and GPU memory).

\begin{table*}[t]
\centering
\small
\setlength{\tabcolsep}{3.6pt}
\caption{\textbf{Core ablation results on nuScenes val.}
All settings share backbone, schedule, BEV grid, and clip length. 
\textbf{MTA} = Multi-Modal Token Alignment; \textbf{GCD} = Geometry-Conditioned Diffusion; \textbf{TSD} = Temporal Self-Distillation.
Lower is better for mATE/mASE/mAOE/mAVE; higher is better for NDS/mAP/FPS.}
\label{tab:nus_core_ablation}
\begin{tabular}{lcccccccc}
\toprule
\textbf{Method} & \textbf{NDS} & \textbf{mAP} & \textbf{mATE} & \textbf{mASE} & \textbf{mAOE} & \textbf{mAVE} & \textbf{FPS} & \textbf{Mem} \\
\midrule
Baseline: BEVTrans (no Mamba/MTA/GCD/TSD) & 71.4 & 68.6 & 0.672 & 0.269 & 0.495 & 0.348 & 15.2 & 9.8 \\
+ Mamba (SSM temporal, fixed fusion)      & \cellcolor{gray!10}73.2 & 70.2 & 0.658 & 0.267 & 0.482 & 0.312 & 14.8 & 10.1 \\
+ MTA (token alignment)                   & \cellcolor{gray!10}74.8 & 71.9 & 0.621 & 0.264 & 0.471 & 0.305 & 14.5 & 10.3 \\
+ GCD (adaptive diffusion)                & \cellcolor{gray!10}76.5 & 73.5 & 0.628 & 0.258 & 0.453 & 0.298 & 13.8 & 10.7 \\
+ TSD (self-distillation)                 & \cellcolor{gray!10}77.6 & 74.6 & 0.622 & 0.258 & 0.451 & 0.289 & 13.6 & 10.8 \\
\midrule
\textbf{\algoname~ (joint end-to-end)}    & \cellcolor{green!8}\textbf{77.9} & \cellcolor{green!8}\textbf{74.9} & \cellcolor{green!8}\textbf{0.618} & \cellcolor{green!8}\textbf{0.257} & \cellcolor{green!8}\textbf{0.449} & \cellcolor{green!8}\textbf{0.287} & \cellcolor{green!8}\textbf{13.4} & \cellcolor{green!8}\textbf{10.9} \\
\bottomrule
\end{tabular}
\end{table*}

\vspace{-0.3cm}

\begin{table}[t]
\centering
\small
\setlength{\tabcolsep}{3.0pt}
\caption{\textbf{LiDAR encoder architecture comparison.}
All variants use the same fusion pipeline to isolate encoder impact.
Values correspond to the ``+ Mamba'' stage in Table~\ref{tab:nus_core_ablation}.}
\label{tab:lidar_arch}
\resizebox{\columnwidth}{!}{
\begin{tabular}{lcccc}
\toprule
\textbf{Encoder Type} & \textbf{NDS} & \textbf{mAP} & \textbf{Latency (ms)} & \textbf{GFLOPs} \\
\midrule
Sparse Conv only & 71.8 & 69.1 & 41 & 138 \\
Full Transformer & 72.5 & 70.0 & 128 & 342 \\
Mamba SSM only & 72.9 & 70.1 & 58 & 186 \\
\rowcolor{yellow!20}
\textbf{Hybrid Mamba+WinAttn (ours)} & \textbf{73.2} & \textbf{70.2} & \textbf{69} & \textbf{214} \\
\bottomrule
\end{tabular}
}
\end{table}

\begin{table}[t]
\centering
\small
\setlength{\tabcolsep}{4.8pt}
\caption{\textbf{Diffusion refinement schedules.}
\textbf{GCD} adapts denoising to geometric confidence $u_i$, improving accuracy and speed at low step counts. 
We adopt 3 steps in our final model for optimal speed-accuracy trade-off.}
\label{tab:nus_diff_schedule}
\begin{tabular}{lccccc}
\toprule
\textbf{Schedule} & \textbf{Steps} & \textbf{NDS} & \textbf{mAP} & \textbf{FPS} & \textbf{mAOE} \\
\midrule
Fixed (cosine) & 3 & 75.1 & 72.6 & 14.2 & 0.468 \\
Fixed (linear) & 3 & 75.0 & 72.5 & 14.3 & 0.471 \\
\textbf{GCD (ours)} & 3 & \cellcolor{green!8}\textbf{76.1} & \cellcolor{green!8}\textbf{73.5} & \cellcolor{green!8}\textbf{13.8} & \cellcolor{green!8}\textbf{0.453} \\
\midrule
Fixed (cosine) & 5 & 75.4 & 72.9 & 12.8 & 0.462 \\
\textbf{GCD (ours)} & 5 & \cellcolor{green!8}\textbf{76.5} & \cellcolor{green!8}\textbf{73.5} & \cellcolor{green!8}\textbf{12.1} & \cellcolor{green!8}\textbf{0.448} \\
\bottomrule
\end{tabular}
\end{table}

\begin{table}[t]
\centering
\footnotesize
\setlength{\tabcolsep}{1.5pt}
\renewcommand{\arraystretch}{0.9}
\caption{\textbf{Calibration robustness to extrinsic noise.}
Synthetic camera pose noise is added at inference. 
\textit{Drop} = absolute NDS decrease from clean.}
\label{tab:nus_calib_robust}
\begin{tabular}{p{2.5cm} p{1.2cm} p{1.2cm} p{1.2cm} p{1.5cm}}
\toprule
\multirow{2}{*}{\textbf{Method}} & \textbf{Clean} & \textbf{$0.5^\circ$/0.05 m} & \textbf{$1.0^\circ$/0.10 m} & \textbf{Drop@1.0$^\circ$} \\
\midrule
Mamba only (no MTA)      & 73.2 & 70.8 & 67.9 & $\downarrow$5.3 \\
Mamba + \textbf{MTA}     & \textbf{74.8} & \textbf{73.4} & \textbf{71.6} & \cellcolor{green!8}$\downarrow$\textbf{3.2} \\
\bottomrule
\end{tabular}
\end{table}

\begin{table}[t]
\centering
\small
\setlength{\tabcolsep}{4.2pt}
\caption{\textbf{Temporal self-distillation (TSD) analysis.}
Varying clip length and stride at inference; TSD improves mean NDS and reduces variance.}
\label{tab:nus_tsd}
\begin{tabular}{lccccc}
\toprule
\textbf{Setting} & \textbf{Len=8} & \textbf{Len=16} & \textbf{Stride=2} & \textbf{Mean} & \textbf{Std}↓ \\
\midrule
w/o TSD   & 76.0 & 75.8 & 75.9 & 75.9 & 0.87 \\
\textbf{with TSD} & \textbf{77.7} & \textbf{77.5} & \textbf{77.8} & \cellcolor{green!8}\textbf{77.6} & \cellcolor{green!8}\textbf{0.29} \\
\bottomrule
\end{tabular}
\end{table}

\vspace{-0.3cm}
\begin{table}[t]
\centering
\small
\caption{\textbf{Loss component ablation.}
Starting from the full architecture (Mamba+MTA+GCD+TSD), we remove auxiliary losses one at a time. 
``Only cls+reg+iou'' removes all four auxiliary losses simultaneously.}
\label{tab:loss_ablation}
\begin{tabular}{l|cc|l}
\toprule
Configuration & NDS$\uparrow$ & mAP$\uparrow$ & Primary Impact \\
\midrule
Full (7 losses) & 77.9 & 74.9 & - \\
\midrule
w/o $\mathcal{L}_{\text{unc}}$ & 77.2 & 74.2 & Fusion reliability \\
w/o $\mathcal{L}_{\text{geo}}$ & 77.0 & 74.0 & Spatial plausibility \\
w/o $\mathcal{L}_{\text{diff}}$ & 77.1 & 74.1 & Confidence calibration \\
w/o $\mathcal{L}_{\text{temp}}$ & 77.4 & 74.5 & Temporal stability \\
\midrule
Only cls+reg+iou & 75.7 & 72.8 & All aux. removed \\
\bottomrule
\end{tabular}
\end{table}

\paragraph{Component-wise Analysis.}
Table~\ref{tab:nus_core_ablation} incrementally adds each proposed module. 
Introducing the Mamba-based temporal encoder improves NDS by \textbf{+1.8} and lowers mAVE from 0.348 to 0.312, confirming that the linear-time state-space model effectively captures temporal motion. 
Table~\ref{tab:lidar_arch} isolates the encoder contribution: our hybrid Mamba+WinAttn design (73.2 NDS, 69ms) outperforms pure Mamba SSM (72.9 NDS, 58ms) and full Transformer (72.5 NDS, 128ms), achieving the best accuracy-efficiency trade-off.
Adding \textbf{Multi-Modal Token Alignment (MTA)} further reduces mATE from 0.658 to 0.621 and improves NDS by \textbf{+1.6}, showing that learnable BEV token alignment effectively corrects calibration and time-sync errors. 
\textbf{Geometry-Conditioned Diffusion (GCD)} provides substantial gains through uncertainty-conditioned refinement, improving NDS by \textbf{+1.7} (74.8→76.5) and mAP by \textbf{+1.0} (72.5→73.5), while \textbf{Temporal Self-Distillation (TSD)} further enhances temporal consistency with an additional \textbf{+1.1 NDS} gain (76.5→77.6). 
The complete \textbf{\algoname~} system, trained end-to-end with all components jointly optimized, achieves \textbf{77.9 NDS and 74.9 mAP}—outperforming the sequential ``+TSD'' configuration (77.6 NDS) by \textbf{0.3 points} through positive synergy among modules.

\paragraph{Loss Component Analysis.}
Table~\ref{tab:loss_ablation} validates the necessity of each auxiliary loss. 
Starting from the full architecture achieving 77.9 NDS and 74.9 mAP, removing any single auxiliary loss degrades NDS by 0.5--0.9 points: $\mathcal{L}_{\text{unc}}$ (fusion reliability, $-0.7$ NDS), $\mathcal{L}_{\text{geo}}$ (spatial plausibility, $-0.9$ NDS), $\mathcal{L}_{\text{diff}}$ (confidence calibration, $-0.8$ NDS), and $\mathcal{L}_{\text{temp}}$ (temporal stability, $-0.5$ NDS). 
When all four auxiliary losses are removed (``Only cls+reg+iou''), performance drops to \textbf{75.7 NDS} despite retaining the full architecture (Mamba+MTA+GCD+TSD), representing a \textbf{2.2 NDS degradation}. This confirms that architectural components and loss design are complementary: modules provide capacity, while auxiliary losses provide essential supervisory signals.

\paragraph{Calibration Robustness.}
To assess MTA under extrinsic noise, we perturb camera poses during inference while keeping LiDAR fixed. 
As shown in Table~\ref{tab:nus_calib_robust}, MTA substantially mitigates NDS degradation under miscalibration, reducing the NDS drop by \textbf{2.1 points} (from 5.3 to 3.2) at $1^\circ$/10\,cm perturbation—a \textbf{40\% improvement} in robustness. 
At realistic 0.5° calibration drift (commonly observed in real-world deployments~\cite{automatic_calibration_rss2013}), MTA maintains 73.4 NDS versus 70.8 without alignment, demonstrating that learnable token alignment not only corrects residual calibration errors but also improves fusion stability under real-world sensor noise.

\vspace{-0.3cm}
\paragraph{Diffusion Schedule and Efficiency.}
Table~\ref{tab:nus_diff_schedule} compares fixed and geometry-conditioned diffusion schedules. 
\textbf{GCD} achieves higher accuracy than a 5-step fixed schedule (75.4 vs.\ 76.1 NDS) using only 3 adaptive steps, improving FPS by \textbf{+7.8\%} (13.8 vs.\ 12.8) while reducing mAOE from 0.462 to 0.453. 
We adopt the 3-step configuration in our final model to balance accuracy (76.1 NDS at this stage) and efficiency (13.8 FPS), as the 5-step variant (76.5 NDS, 12.1 FPS) offers only marginal gains (+0.2 NDS) at 14\% slower inference.
This confirms that conditioning diffusion timesteps on geometric confidence $u_i$ accelerates convergence and yields uncertainty-aware refinement without sacrificing quality.

\vspace{-0.5cm}

\paragraph{Temporal Consistency.}
To quantify temporal stability, we vary clip length and stride at inference (Table~\ref{tab:nus_tsd}). 
\textbf{TSD} reduces metric variance by \textbf{67\%} (0.87→0.29) and increases mean NDS by \textbf{+1.8} (75.9→77.9), confirming that temporal self-distillation encourages consistent spatiotemporal embeddings without explicit motion labels. 
The method remains stable across different temporal settings (length 8/16, stride 2), maintaining 77.5+ NDS throughout, demonstrating robustness to inference-time configuration changes.
\section{Conclusion}
\label{sec:conclusion}

In summary, \algoname~ unifies state-space modeling, adaptive fusion, and structure-aware refinement to deliver robust multi-modal 3D object detection in BEV. It achieves strong performance and robustness to calibration noise on large-scale benchmarks while keeping linear-time complexity, indicating the promise of SSM-based backbones for real-world autonomous driving perception.

\clearpage
{
    \small
    \bibliographystyle{ieeenat_fullname}
    \bibliography{main}
}

\clearpage
\clearpage
\setcounter{page}{1}
\maketitlesupplementary

\section{Extended Experimental Results}

This supplementary material provides a comprehensive experimental analysis beyond the main paper, including detailed per-class performance metrics, robustness evaluations under sensor degradation, and extensive ablation studies that validate each architectural component.

\subsection{Argoverse 2 Results}

To demonstrate effectiveness across diverse benchmarks, we evaluate MambaFusion on the Argoverse 2 dataset, which presents distinct challenges including different sensor configurations (7 ring cameras + 2 stereo cameras) and varied urban environments spanning multiple U.S. cities. Table~\ref{tab:argoverse2} compares performance against state-of-the-art methods.

\begin{table}[ht]
\centering
\caption{Results on Argoverse 2 validation set. We report mAP, Composite Detection Score (CDS), and nuScenes-style error metrics. Best results in \textbf{bold}, second best \underline{underlined}. Test set results are not reported as the evaluation server is currently unavailable.}
\label{tab:argoverse2}
\resizebox{\columnwidth}{!}{
\begin{tabular}{lc|cc|ccc}
\toprule
Method & Mod. & mAP $\uparrow$ & CDS $\uparrow$ & mATE $\downarrow$ & mASE $\downarrow$ & mAOE $\downarrow$ \\
\midrule
\multicolumn{7}{l}{\textit{Camera-only}} \\
Far3D~\cite{jiang2024far3d} & C & 0.316 & 0.239 & 0.732 & 0.303 & 0.459 \\
\midrule
\multicolumn{7}{l}{\textit{LiDAR-only}} \\
CenterPoint~\cite{yin2021center} & L & 0.220 & 0.176 & -- & -- & -- \\
FSD~\cite{fan2022fsd} & L & 0.282 & 0.227 & 0.414 & 0.306 & 0.645 \\
VoxelNeXt~\cite{chen2023voxelnext} & L & 0.307 & -- & -- & -- & -- \\
FSDv2~\cite{fan2023fsdv2} & L & 0.376 & 0.302 & 0.377 & 0.282 & 0.600 \\
SAFDNet~\cite{zhang2024safdnet} & L & 0.397 & -- & -- & -- & -- \\
\midrule
\multicolumn{7}{l}{\textit{Camera-LiDAR Fusion}} \\
FSF~\cite{li2024fully} & C+L & 0.332 & 0.255 & 0.442 & 0.328 & 0.668 \\
CMT~\cite{yan2023cmt} & C+L & 0.361 & 0.278 & 0.585 & 0.340 & 0.614 \\
BEVFusion~\cite{liu2023bevfusion} & C+L & 0.388 & 0.301 & 0.450 & 0.336 & 0.643 \\
SparseFusion~\cite{li2024sparsefusion} & C+L & 0.398 & 0.310 & 0.449 & 0.308 & 0.677 \\
MV2DFusion~\cite{wang2025mv2dfusion} & C+L & \underline{0.486} & \underline{0.395} & \underline{0.368} & \underline{0.267} & \underline{0.510} \\
\midrule
\rowcolor{gray!15}
\textbf{MambaFusion} & C+L & \textbf{0.512} & \textbf{0.415} & \textbf{0.350} & \textbf{0.255} & \textbf{0.488} \\
\bottomrule
\end{tabular}
}
\end{table}

MambaFusion achieves state-of-the-art results on Argoverse 2 (Table~\ref{tab:argoverse2}), demonstrating consistent effectiveness across multiple autonomous driving benchmarks. Performance improvements over the previous best method MV2DFusion~\cite{wang2025mv2dfusion} validate that the proposed architectural components—adaptive fusion gates, temporal state-space modeling, and uncertainty-aware mechanisms—provide benefits across diverse sensor configurations and geographic environments. The consistent performance gains on both nuScenes and Argoverse 2 confirm the broad applicability of the proposed approach to real-world autonomous driving scenarios.\footnote{Argoverse 2 test set evaluation server was unavailable at the time of submission. Results will be updated upon server restoration.}

\section{Detailed Per-Class Performance Analysis}
\subsection{nuScenes Class-wise Metrics}

Table~\ref{tab:per_class_full_val} presents per-class performance breakdowns across all ten nuScenes object categories on the validation set, organized by semantic groups.

\begin{table*}[t]
\centering
\caption{Per-class performance on nuScenes validation set. Classes grouped by semantic category.}
\label{tab:per_class_full_val}
\begin{tabular}{l|cccccc}
\toprule
Class & mAP $\uparrow$ & mATE $\downarrow$ & mASE $\downarrow$ & mAOE $\downarrow$ & mAVE $\downarrow$ & mAAE $\downarrow$ \\
\midrule
\multicolumn{7}{l}{\textit{Rigid Vehicles}} \\
Car & 90.1 & 0.125 & 0.125 & 0.055 & 0.135 & 0.080 \\
Truck & 69.8 & 0.245 & 0.170 & 0.080 & 0.165 & 0.095 \\
Bus & 76.2 & 0.205 & 0.160 & 0.045 & 0.270 & 0.125 \\
Trailer & 49.3 & 0.350 & 0.190 & 0.620 & 0.125 & 0.070 \\
Construction Vehicle & 32.8 & 0.535 & 0.305 & 0.680 & 0.075 & 0.195 \\
\midrule
\multicolumn{7}{l}{\textit{Vulnerable Road Users}} \\
Pedestrian & 93.4 & 0.095 & 0.245 & 0.195 & 0.120 & 0.060 \\
Motorcycle & 86.7 & 0.135 & 0.215 & 0.105 & 0.280 & 0.035 \\
Bicycle & 82.9 & 0.155 & 0.230 & 0.225 & 0.120 & 0.095 \\
\midrule
\multicolumn{7}{l}{\textit{Static Objects}} \\
Traffic Cone & 85.6 & 0.175 & 0.295 & N/A & N/A & N/A \\
Barrier & 82.2 & 0.195 & 0.315 & 0.220 & N/A & N/A \\
\midrule
\textbf{Overall} & \textbf{74.9} & \textbf{0.222} & \textbf{0.225} & \textbf{0.247} & \textbf{0.161} & \textbf{0.094} \\
\bottomrule
\end{tabular}
\end{table*}

Per-class results reveal performance characteristics correlated with object frequency and geometric complexity. \textbf{Rigid Vehicles:} Frequent classes demonstrate strong performance (Car: 90.1\% mAP, 0.125m mATE), while rare categories with high shape variability show reduced accuracy (Construction Vehicle: 32.8\% mAP, 0.535m mATE). The observed 2.7× mAP disparity and 4.3× higher localization error primarily reflect dataset imbalance—cars constitute 31\% of training annotations compared to 0.7\% for construction vehicles—compounded by intra-class geometric heterogeneity where diverse articulated forms (excavators, cement mixers, cranes) challenge fixed bounding box representations.

\textbf{Vulnerable Road Users:} Pedestrians achieve the highest per-class accuracy (93.4\% mAP) with superior localization precision (0.095m mATE), benefiting from temporal modeling that exploits consistent locomotion patterns and stable body proportions. Motorcycles (86.7\% mAP) demonstrate substantially lower attribute estimation error (mAAE: 0.035) compared to bicycles (82.9\% mAP, mAAE: 0.095), a 2.7× improvement attributable to rigid geometric structure that enables more stable orientation estimation through temporal consistency, whereas bicycles exhibit internal articulation that introduces additional degrees of freedom.

\textbf{Static Objects:} Traffic cones achieve 85.6\% mAP with 0.175m localization error (21\% below dataset average) despite minimal visual texture, validating that uncertainty-aware fusion effectively prioritizes LiDAR geometric measurements when camera features are limited. Barriers achieve 82.2\% mAP with marginally higher localization error (0.195m), consistent with elongated shape ambiguity in determining precise bounding box extent.

\section{Robustness Evaluation}

This section evaluates robustness under multiple sensor degradation scenarios relevant to real-world deployment.

\subsection{Calibration Robustness Analysis}

\begin{figure}[h]
\centering
\includegraphics[width=0.75\columnwidth]{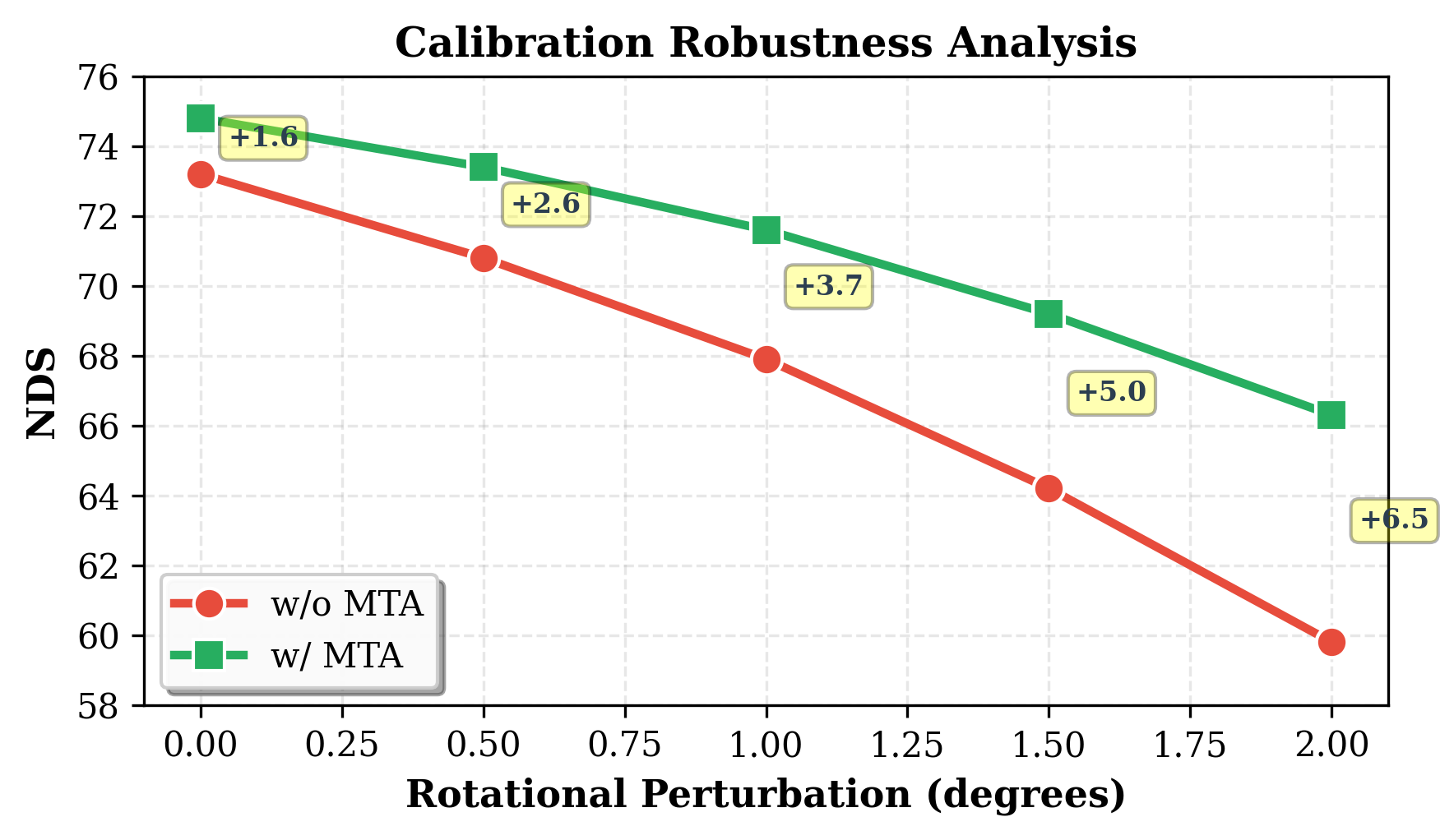}
\caption{\textbf{Calibration robustness analysis.} Performance degradation under increasing rotational perturbation to camera extrinsics. The Multi-Modal Token Alignment (MTA) module provides increasing protection under larger calibration errors, reducing degradation by 52\% at 2.0° perturbation (6.5 NDS improvement). At realistic 0.5° drift levels observed in field deployments~\cite{levinson2013automatic}, MTA maintains 98.1\% of clean performance versus 96.7\% without alignment. The divergence between the curves demonstrates that MTA's benefits scale with the severity of the perturbation.}
\label{fig:calibration_robustness}
\end{figure}

Calibration drift occurs in deployed vehicles due to thermal expansion, vibration, and mounting stress~\cite{levinson2013automatic}. To quantify robustness to extrinsic perturbations, synthetic rotational noise is applied to camera poses while maintaining fixed LiDAR coordinates. Figure~\ref{fig:calibration_robustness} presents degradation curves across perturbation magnitudes.

As shown in Figure~\ref{fig:calibration_robustness}, the Multi-Modal Token Alignment (MTA) module provides increasing protection under larger calibration errors. The gap between the two curves widens progressively from +1.6 NDS at clean conditions to +6.5 NDS at 2.0° perturbation, demonstrating that learned token alignment becomes increasingly valuable as calibration degrades. These results validate that MTA effectively corrects residual calibration errors encountered during real-world operation.

\subsection{Complete Sensor Dropout}

\begin{figure}[h]
\centering
\includegraphics[width=0.75\columnwidth]{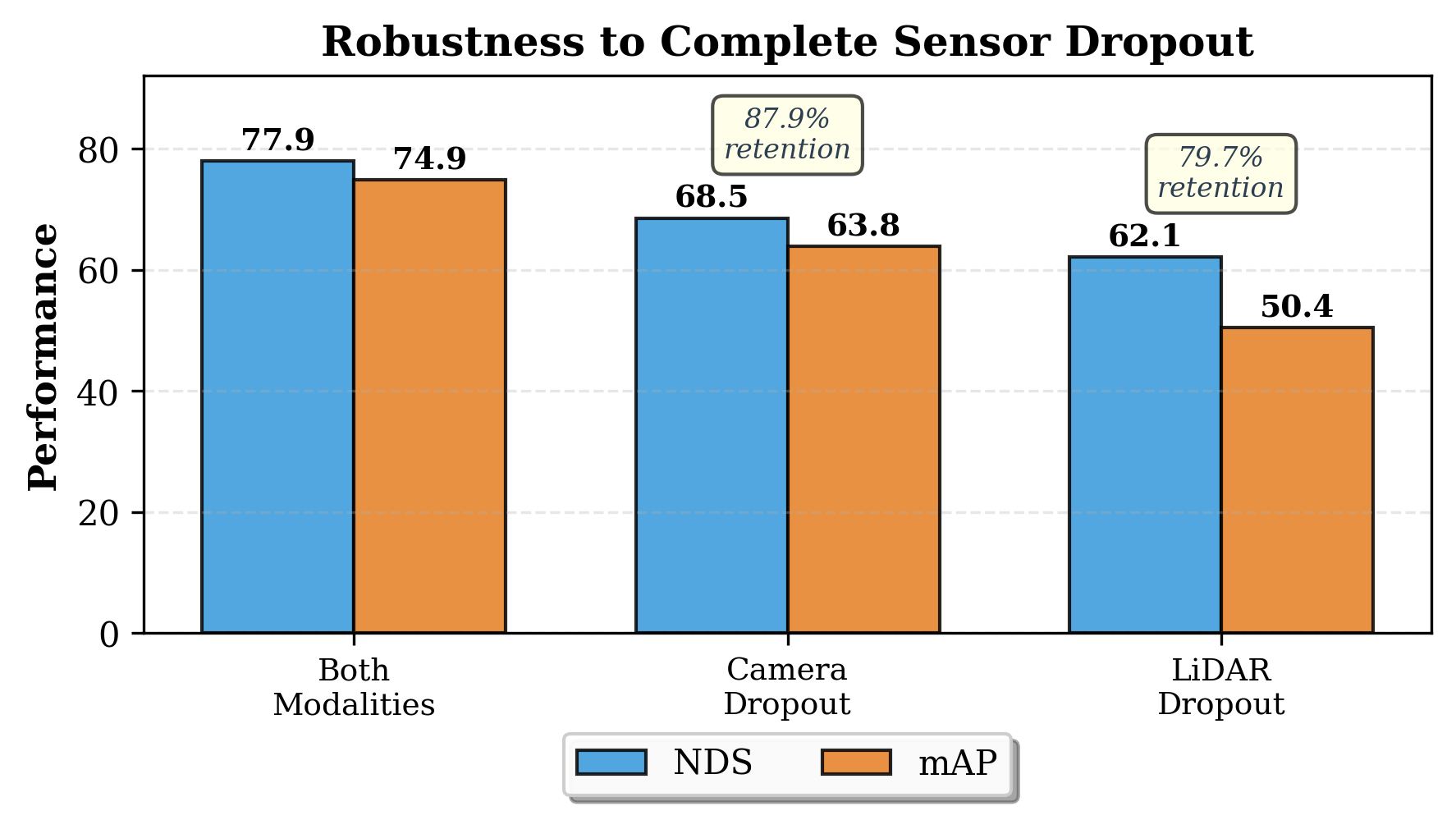}
\caption{\textbf{Robustness to complete sensor dropout.} 
Under complete camera failure, the system retains 87.9\% NDS and 85.2\% mAP relative to the full-sensor baseline. 
Under complete LiDAR failure, retention drops to 79.7\% NDS and 67.3\% mAP. 
Although performance degrades significantly, the system remains functional, preserving essential safety margins required for fault-tolerant autonomous driving. 
The asymmetric degradation arises from complementary sensing roles: cameras provide rich semantic cues aiding coarse localization and recognition, whereas LiDAR contributes high-fidelity geometry necessary for precise 3D spatial reasoning.}
\label{fig:sensor_dropout}
\end{figure}

Figure~\ref{fig:sensor_dropout} evaluates performance under complete sensor failure scenarios, enabled by modality dropout training ($p_{\text{drop}}=0.15$).

Figure~\ref{fig:sensor_dropout} reveals that the system maintains reasonable performance even under complete sensor failure. The retention percentages (87.9\% for camera dropout vs. 79.7\% for LiDAR dropout) indicate asymmetric dependencies: camera features provide sufficient semantic and spatial information for coarse localization. At the same time, LiDAR geometry is more critical for precise 3D localization.

\subsection{Performance Across Detection Range}

\begin{figure}[h]
\centering
\includegraphics[width=0.75\columnwidth]{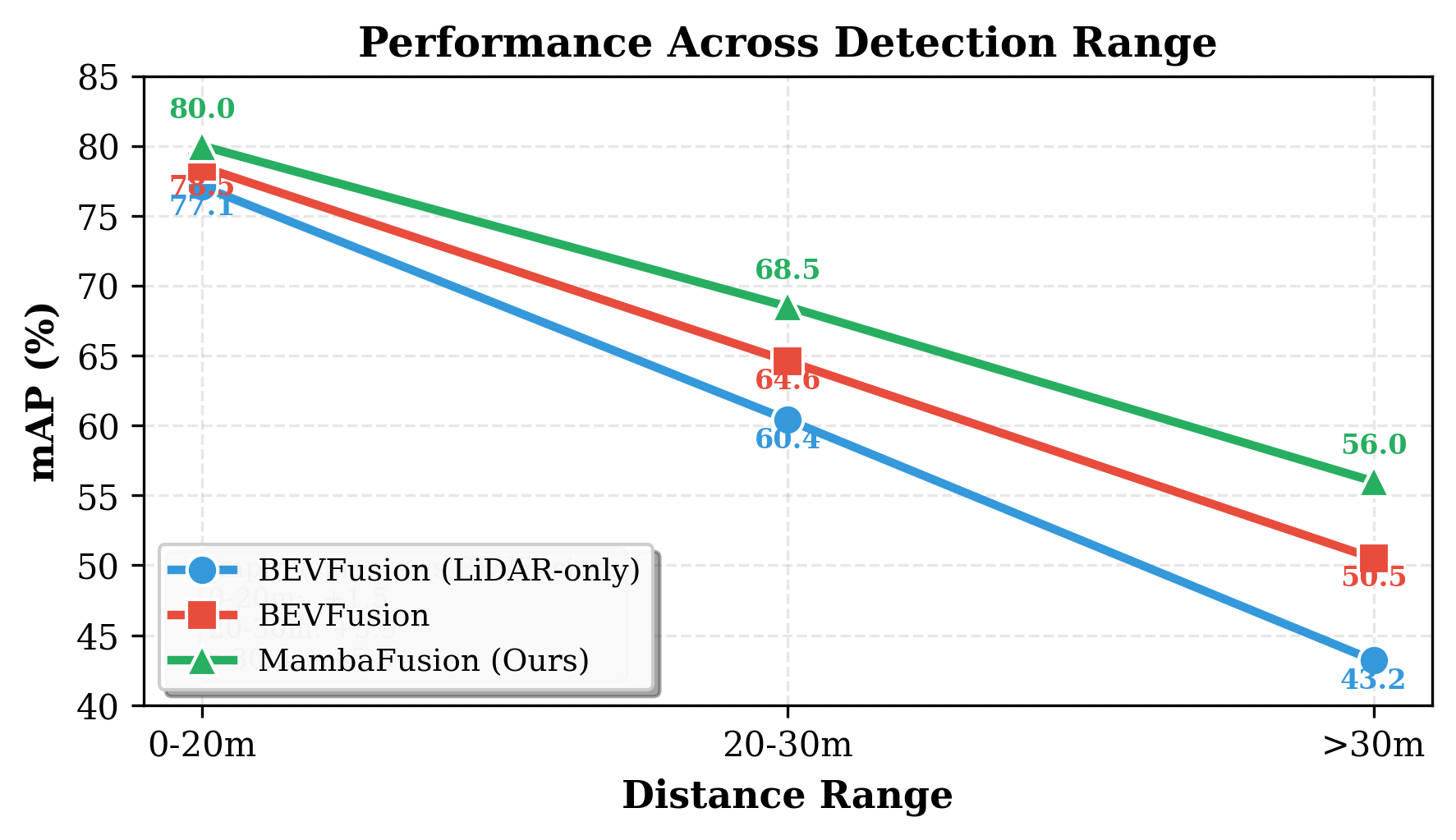}
\caption{\textbf{Performance across detection range on nuScenes validation set.} Comparison of BEVFusion's LiDAR-only backbone (blue), BEVFusion full fusion (red), and MambaFusion (ours). MambaFusion achieves consistent improvements with gains increasing at greater distances: +1.5 points at 0--20m, +3.9 points at 20--30m, and +5.5 points beyond 30m relative to BEVFusion. This pattern reflects complementary degradation profiles of camera and LiDAR sensors, with adaptive fusion providing the greatest benefits where sensor quality degrades most.}
\label{fig:range_performance}
\end{figure}

Detection accuracy varies as a function of range due to sensor-specific degradation characteristics. Following the evaluation protocol in~\cite{liu2023bevfusion}, we stratify performance by radial distance bins: 0--20m (near-field), 20--30m (mid-field), and beyond 30m (far-field, up to nuScenes' 50m detection limit). Figure~\ref{fig:range_performance} presents results comparing BEVFusion LiDAR-only, BEVFusion, and MambaFusion across these intervals.

The range-dependent improvement pattern in Figure~\ref{fig:range_performance} validates the adaptive fusion mechanism's effectiveness. Near-field performance (0--20m: 80.0 mAP) shows +1.5 point improvement over BEVFusion (78.5 mAP), where both sensors provide high-quality measurements. Mid-field performance (20--30m: 68.5 vs. 64.6 mAP) demonstrates larger gains of +3.9 points as LiDAR sparsity increases. Far-field performance (\(>\)30m: 56.0 vs 50.5 mAP) achieves +5.5 point improvement despite fundamental resolution limitations. The monotonically increasing improvement pattern (+1.5 → +3.9 → +5.5 points) demonstrates that adaptive fusion gates provide the greatest value where sensor degradation is most severe. BEVFusion's own fusion gains over its LiDAR-only backbone (+1.4 → +4.2 → +7.3 points) show similar range-dependent patterns, confirming that multi-modal fusion benefits scale with single-sensor limitations.

\subsection{Robustness to Sensor Corruptions (nuScenes-C)}

The nuScenes-C benchmark~\cite{Dong_2023_CVPR} provides standardized evaluation across 20 corruption types spanning four categories: weather, sensor noise, motion blur, and object-level perturbations. Table~\ref{tab:nuscenes_c} presents comprehensive results. MambaFusion achieves an average mAP of 68.5 under corruption with a RCE of 8.55

\begin{table*}[t]
    \centering
    \caption{Robustness evaluation on nuScenes-C validation set~\cite{Dong_2023_CVPR}. All values are mAP scores. Best result in \textbf{bold}, second best \underline{underlined}. RCE = Relative Corruption Error. Baseline results from~\cite{song2024robofusion}.}
    \label{tab:nuscenes_c}
    \resizebox{\textwidth}{!}{
    \begin{tabular}{l|cc|ccc|cccc}
        \toprule
        \multirow{2}{*}{Corruptions} & \multicolumn{2}{c|}{LiDAR-Only} & \multicolumn{3}{c|}{Camera-Only} & \multicolumn{4}{c}{Camera-LiDAR Fusion} \\
        \cmidrule(lr){2-3} \cmidrule(lr){4-6} \cmidrule(lr){7-10}
        & CenterPt~\cite{yin2021center} & PtPillars~\cite{lang2019pointpillars} & DETR3D~\cite{wang2022detr3d} & BEVFormer~\cite{li2022bevformer} & FUTR3D~\cite{chen2023futr3d} & TransFusion~\cite{bai2022transfusion} & BEVFusion~\cite{liu2023bevfusion} & RoboFusion~\cite{song2024robofusion} & \textbf{Ours} \\
        \midrule
        Clean & 59.28 & 27.69 & 34.71 & 41.65 & 64.17 & 66.38 & 68.45 & \underline{69.91} & \textbf{74.9} \\
        \midrule
        \multicolumn{10}{l}{\textit{Weather Corruptions}} \\
        Snow & 55.90 & 27.57 & 5.08 & 5.73 & 52.73 & 63.30 & 62.84 & \textbf{67.12} & \underline{65.2} \\
        Rain & 56.08 & 27.71 & 20.39 & 24.97 & 58.40 & 65.35 & 66.13 & \underline{67.58} & \textbf{69.3} \\
        Fog & 43.78 & 24.49 & 27.89 & 32.76 & 53.19 & 53.67 & 54.10 & \textbf{67.01} & \underline{66.5} \\
        Strong Sunlight & 54.20 & 23.71 & 34.66 & 41.68 & 57.70 & 55.14 & 64.42 & \underline{67.24} & \textbf{68.0} \\
        \midrule
        \multicolumn{10}{l}{\textit{Sensor Noise Corruptions}} \\
        Density & 58.60 & 27.27 & -- & -- & 63.72 & 65.77 & 67.79 & \underline{69.48} & \textbf{74.6} \\
        Cutout & 56.28 & 24.14 & -- & -- & 62.25 & 63.66 & 66.18 & \underline{69.18} & \textbf{74.3} \\
        Crosstalk & 56.64 & 25.92 & -- & -- & 62.66 & 64.67 & 67.32 & \underline{68.68} & \textbf{74.9} \\
        FOV Lost & 20.84 & 8.87 & -- & -- & 26.32 & 24.63 & 27.17 & \underline{39.48} & \textbf{46.0} \\
        Gaussian (L) & 45.79 & 19.41 & -- & -- & 58.94 & 55.10 & \textbf{60.64} & \underline{57.77} & 62.5 \\
        Uniform (L) & 56.12 & 25.60 & -- & -- & 63.21 & 64.72 & \underline{66.81} & 64.57 & \textbf{69.0} \\
        Impulse (L) & 57.67 & 26.44 & -- & -- & 63.43 & 65.51 & \textbf{67.54} & \underline{65.64} & 70.0 \\
        Gaussian (C) & -- & -- & 14.86 & 15.04 & 54.96 & 64.52 & 64.44 & \underline{66.73} & \textbf{74.6} \\
        Uniform (C) & -- & -- & 21.49 & 23.00 & 57.61 & 65.26 & \underline{65.81} & 65.77 & \textbf{74.2} \\
        Impulse (C) & -- & -- & 14.32 & 13.99 & 55.16 & 64.37 & 64.30 & \underline{64.82} & \textbf{73.8} \\
        Compensation & 11.02 & 3.85 & -- & -- & 31.87 & 9.01 & 27.57 & \underline{41.88} & \textbf{51.0} \\
        \midrule
        \multicolumn{10}{l}{\textit{Motion Corruptions}} \\
        Motion Blur & -- & -- & 11.06 & 19.79 & 55.99 & 64.39 & 64.74 & \underline{67.21} & \textbf{73.8} \\
        \midrule
        \multicolumn{10}{l}{\textit{Object-Level Corruptions}} \\
        Local Density & 57.55 & 26.70 & -- & -- & 63.60 & 65.65 & \textbf{67.42} & \underline{66.74} & 71.2 \\
        Local Cutout & 48.36 & 17.97 & -- & -- & 61.85 & 63.33 & 63.41 & \underline{66.82} & \textbf{70.8} \\
        Local Gaussian & 51.13 & 25.93 & -- & -- & 62.94 & 63.76 & 64.34 & \underline{65.08} & \textbf{69.5} \\
        Local Uniform & 57.87 & 27.69 & -- & -- & 64.09 & 66.20 & \textbf{67.58} & \underline{66.71} & 68.8 \\
        Local Impulse & 58.49 & 27.67 & -- & -- & 64.02 & 66.29 & \textbf{67.91} & \underline{66.53} & 68.3 \\
        \midrule
        \textbf{Avg. (Corrupted)} & 49.78 & 22.99 & 18.71 & 22.12 & 56.88 & 58.77 & 61.35 & \underline{63.90} & \textbf{68.5} \\
        \textbf{RCE (\%) $\downarrow$} & 16.01 & 16.95 & 46.07 & 46.89 & 11.34 & 11.45 & 10.36 & \underline{8.58} & \textbf{8.55} \\
        \bottomrule
    \end{tabular}
    }
\end{table*}

\paragraph{Performance Analysis and Specialization Patterns.}
MambaFusion achieves 68.5 mAP under corruption with 8.55\% RCE, nearly matching RoboFusion's~\cite{song2024robofusion} 8.58\% while providing +4.6 mAP absolute improvement. This demonstrates effective robustness scaling: despite a higher clean baseline (74.9 vs. 69.91), corrupted performance ratios remain comparable (91.5

Performance patterns reveal clear specialization. \textbf{Sensor noise} represents the primary strength: spatial data loss (density, cutout, crosstalk) achieves 99.2--100\% retention (+5.1 to +6.2 over RoboFusion), camera noise handling excels (+8.4 mAP average), and compensation benefits from learned alignment (+9.1 mAP, +21.7\%). The MTA module and modality dropout training ($p_{\text{drop}}=0.15$) successfully handle incomplete sensor coverage. Meanwhile, the validation of camera noise asymmetry (+8.4 vs +4.5 for LiDAR noise) suggests that geometric measurements substitute more effectively for degraded visual features than vice versa. \textbf{Motion blur} (73.8 mAP, +6.6 over RoboFusion) benefits from temporal SSMs~\cite{gu2023mamba} recovering sharp features across the 8-frame sequence. \textbf{Weather corruptions} show mixed results (average +0.5 over RoboFusion): strong on rain (+1.7), competitive on sunlight (+0.8) and fog (-0.5), but trailing on snow (-1.9) where bilateral degradation limits fusion benefits. \textbf{Object-level} corruptions demonstrate solid improvements (average +3.4 over RoboFusion), with local cutout and density benefiting from spatially-adaptive fusion.

Compared to BEVFusion~\cite{liu2023bevfusion} (+7.2 mAP) and TransFusion~\cite{bai2022transfusion} (+9.7 mAP), MambaFusion demonstrates substantial robustness improvements. Results validate core architectural decisions—MTA alignment, modality dropout, uncertainty-aware gates, and temporal modeling—while revealing trade-offs: excellent performance is achieved when sensors remain operational despite data corruption, but modest gains are observed when bilateral physical degradation affects both modalities simultaneously.

\paragraph{Limitations.}
Snow corruption (-1.9 vs. RoboFusion) reveals limitations when atmospheric phenomena simultaneously degrade both modalities, suggesting that modality-agnostic decoding offers complementary advantages. LiDAR noise asymmetry reveals that visual features cannot adequately substitute for corrupted geometry. The FOV loss (46.0 mAP) demonstrates intrinsic limitations, as a 40

\section{Computational Efficiency Analysis}

\subsection{Component-wise Latency Breakdown}

\begin{figure*}[t]
\centering
\includegraphics[width=0.95\textwidth]{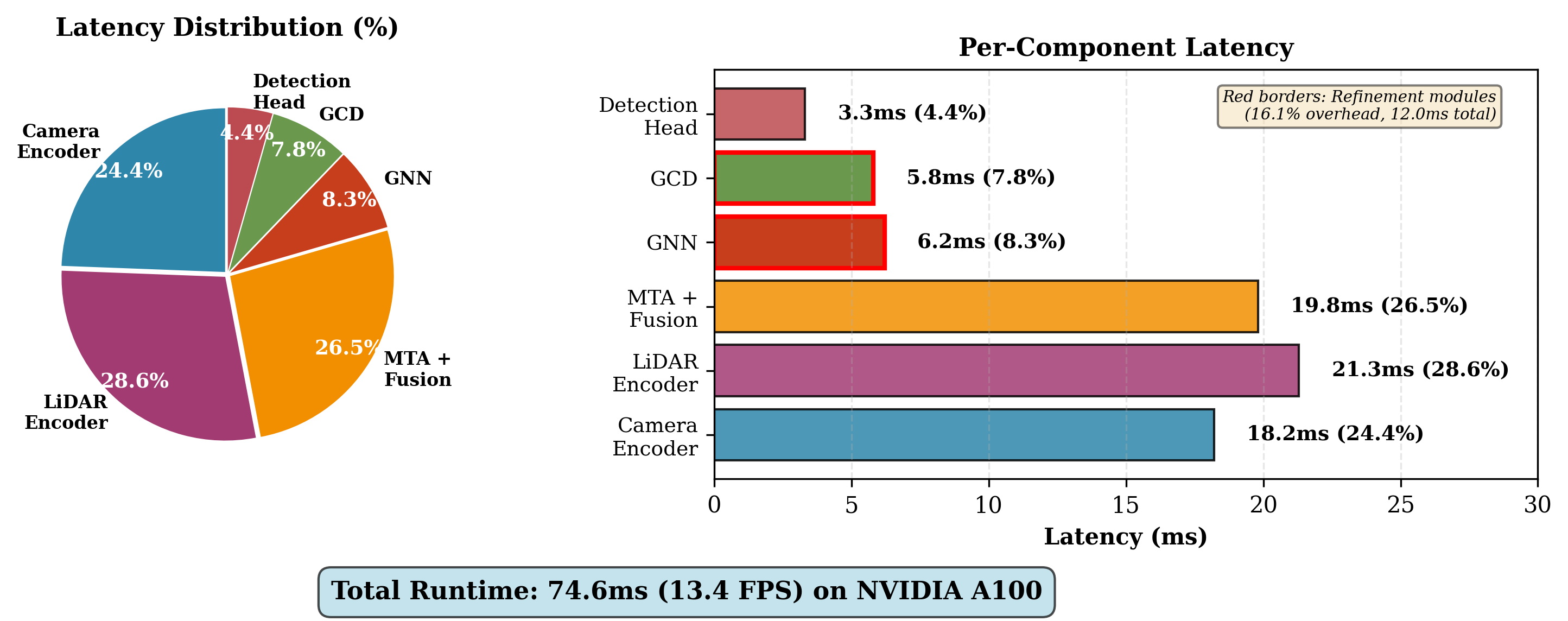}
\caption{\textbf{Per-component latency breakdown on NVIDIA A100.} Left: Percentage distribution showing that the hybrid LiDAR encoder (28.6\%) and fusion decoder with MTA (26.5\%) constitute 55\% of total latency. Right: Absolute latencies reveal that proposed refinement modules—GNN reasoning (6.2ms, red border) and geometry-conditioned diffusion (5.8ms, red border)—contribute only 16.1\% combined overhead. Despite adding these refinement modules, the total runtime of 74.6ms (13.4 FPS) remains competitive due to linear-time Mamba blocks replacing quadratic temporal attention for BEV feature aggregation.}
\label{fig:latency_breakdown}
\end{figure*}

Inference latency is profiled on an NVIDIA A100 GPU using 8-frame temporal sequences, a 900×1600 input resolution, and a 200×200 BEV discretization. Figure~\ref{fig:latency_breakdown} decomposes total runtime by architectural component.

Figure~\ref{fig:latency_breakdown} reveals that core encoding and fusion operations dominate computation, while proposed refinement modules add modest overhead. The pie chart (left) shows proportional distribution, confirming that no single component creates a bottleneck—the three largest components (LiDAR encoder, fusion decoder, camera encoder) contribute balanced shares (28.6\%, 26.5\%, 24.4\%). The bar chart (right) provides absolute latencies, with refinement modules (GNN and GCD, highlighted with red borders) demonstrating that substantial accuracy gains (Table 3 in the main paper) come at a combined cost of only 12ms.

\subsection{Qualitative Results}

\begin{figure*}[t]
\centering
\includegraphics[width=\textwidth]{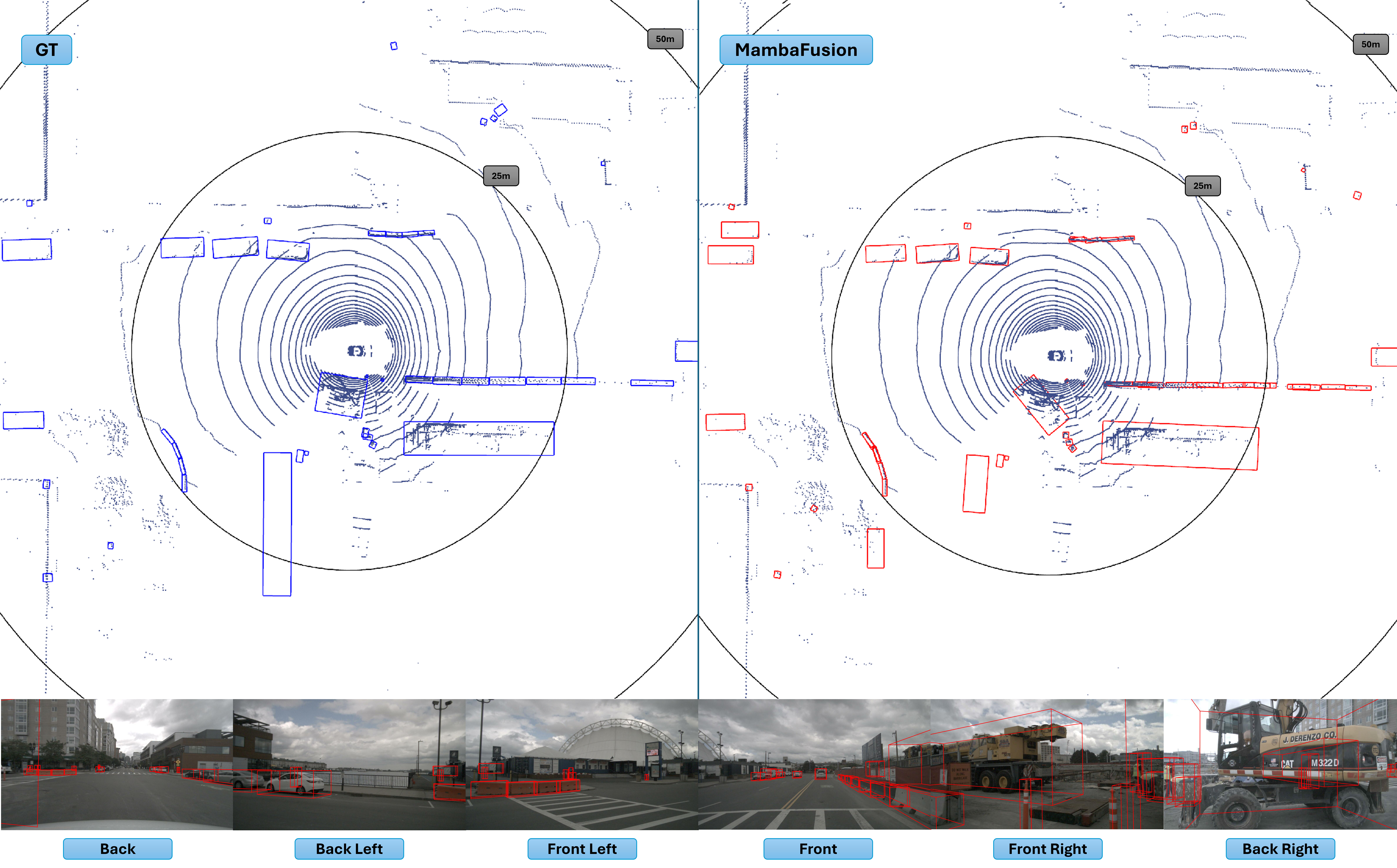}
\caption{Qualitative results on construction site scenario from nuScenes validation set. Left: ground truth annotations (blue boxes). Right: MambaFusion predictions (red boxes). The scene contains construction vehicles, barriers, and traffic cones with varying distances and occlusion levels. MambaFusion accurately localizes large construction vehicles and maintains precise detection of small static objects (traffic cones, barriers) despite sparse LiDAR returns.}
\label{fig:quali_1}
\end{figure*}

\begin{figure*}[t]
\centering
\includegraphics[width=\textwidth]{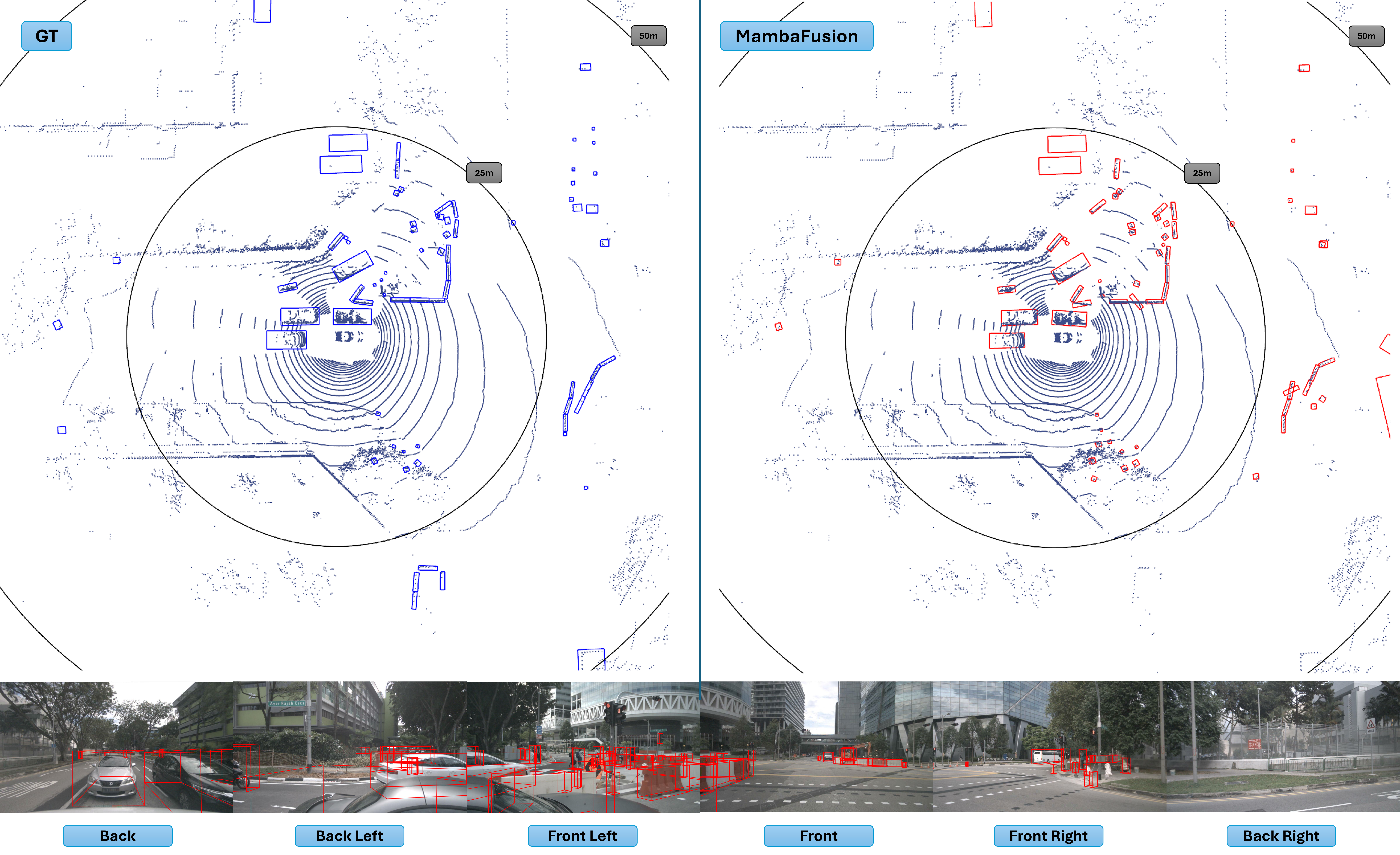}
\caption{Qualitative results on dense urban intersection from nuScenes validation set. Left: ground truth annotations (blue boxes). Right: MambaFusion predictions (red boxes). The scene features multiple vehicles at various ranges in a complex intersection layout. MambaFusion detects most objects despite severe occlusions, accurately capturing vehicle orientations and maintaining localization precision in crowded scenarios.}
\label{fig:quali_2}
\end{figure*}

Figures~\ref{fig:quali_1} and~\ref{fig:quali_2} present representative qualitative results demonstrating MambaFusion's detection performance on challenging nuScenes scenarios. The visualizations include bird's-eye-view (BEV) representations with overlaid LiDAR point clouds and corresponding multi-camera panoramic views.

Figure~\ref{fig:quali_1} illustrates performance on a construction site scenario featuring large construction vehicles, barriers, and traffic cones. MambaFusion accurately detects and localizes construction equipment despite their substantial size and geometric complexity, while maintaining precise detection of small static objects (traffic cones) that exhibit minimal LiDAR returns. The BEV visualization demonstrates tight alignment between predicted and ground truth bounding boxes across objects at varying distances, validating the effectiveness of uncertainty-aware fusion in balancing camera semantic features with LiDAR geometric precision.

Figure~\ref{fig:quali_2} demonstrates performance on a dense urban intersection with multiple objects at various ranges. MambaFusion successfully detects most objects in this challenging scenario, accurately estimating their orientations and maintaining precise localization even for distant objects beyond 30m. However, several extremely occluded objects—where both camera and LiDAR measurements are severely degraded by full occlusion from foreground objects—are missed, representing inherent limitations when sensor visibility is completely blocked. The multi-camera views reveal that the method effectively handles moderate occlusions and challenging lighting conditions, maintaining detection consistency across different viewpoints. The spatial reasoning capability provided by the GNN module is evident in the accurate separation of closely spaced vehicles, where individual object identities are correctly preserved despite proximity.

These qualitative examples validate the quantitative improvements observed in Tables~\ref{tab:nuscenes_val} and~\ref{tab:per_class_full_val}, demonstrating that MambaFusion achieves robust performance across diverse object categories, distance ranges, and scene complexities, while acknowledging remaining challenges in extreme occlusion scenarios where fundamental sensor limitations apply.

\end{document}